\documentclass[11pt]{article}

\usepackage[]{ACL2023}

\usepackage{times}
\usepackage{latexsym}

\usepackage[T1]{fontenc}

\usepackage[utf8]{inputenc}

\usepackage{microtype}

\usepackage{inconsolata}

\usepackage{xspace}
\pdfoutput=1
\usepackage{colortbl}
\usepackage{amsmath}
\usepackage{amsfonts}
\usepackage{graphicx}
\usepackage{ctable}
\usepackage{comment}
\usepackage{multicol}
\usepackage{multirow}
\usepackage{epsfig}
\usepackage{pifont}
\usepackage{dsfont}
\usepackage{makecell}
\usepackage{adjustbox}
\usepackage{txfonts}
\usepackage{xcolor}
\usepackage{soul}
\usepackage{caption}
\usepackage{subcaption}

\title{\system: A Benchmark for Multi-Turn Spoken\\ 
 Conversational Transcript Cleanup}

\newcommand{\system}{\textsc{MultiTurnCleanup}\xspace}

\newcommand{\task}{Multi-Turn Cleanup task\xspace}

\newcommand{\std}{\textsc{STD}\xspace}
\newcommand{\mtd}{\textsc{MTD}\xspace}

\renewcommand{\textsuperscript}[1]{\raisebox{0.5ex}{#1}}

\author{
    Hua Shen\textsuperscript{${\varheartsuit \vardiamondsuit}$\Thanks{ This work was done when the first author was a research intern at Google Research.}}
\quad Vicky Zayats\textsuperscript{${\vardiamondsuit}$}
\quad Johann C. Rocholl\textsuperscript{${\vardiamondsuit}$}
\quad Daniel D. Walker\textsuperscript{${\vardiamondsuit}$}
\quad  Dirk Padfield\textsuperscript{${\vardiamondsuit}$}\\ 
    \textsuperscript{${\varheartsuit}$}University of Michigan,
    \textsuperscript{${\vardiamondsuit}$}Google Research\\
    {\tt huashen@umich.edu}\\
    {\tt \{vzayats,jcrocholl,danwalkeriv,padfield\}@google.com}\\
  }

\begin{document}

\vspace{-1pc}
\maketitle
\vspace{-5pc}

\newcount\Comments  
\Comments=1   

\newcommand{\kibitz}[2]{\ifnum\Comments=1 \textcolor{#1}{#2}\fi}

\newcommand{\todo}[1]{\kibitz{red}{\bf [#1]}}
\newcommand{\hua}[1]{\kibitz{purple}{\bf [#1 -Hua]}}
\newcommand{\vicky}[1]{\kibitz{orange}{\bf [#1 -Vicky]}}
\newcommand{\johann}[1]{\kibitz{blue}{\bf [#1 -Johann]}}
\newcommand{\dan}[1]{\kibitz{green}{\bf [#1 -Dan]}}
\newcommand{\dirk}[1]{\kibitz{cyan}{\bf [#1 -Dirk]}}

\newcommand{\eg}{\emph{e.g.,}\xspace}
\newcommand{\ie}{\emph{i.e.,}\xspace}

\newcommand{\spka}{\textbf{\texttt{A}}\xspace}
\newcommand{\spkb}{\textbf{\texttt{B}}\xspace}

\definecolor{ack-bc}{HTML}{C9DAF8}
\definecolor{rep-bc}{HTML}{F4CCCC}
\definecolor{think-bc}{HTML}{FCE5CD}
\definecolor{inc-bc}{HTML}{D9EAD3}
\definecolor{other-bc}{HTML}{D9D2E9}




\begin{abstract}
Current disfluency detection models focus on individual utterances each from a single speaker. However, numerous discontinuity phenomena in spoken conversational transcripts occur across multiple turns,
which can not be identified by disfluency detection models.
This study addresses these phenomena by proposing an innovative \task for spoken conversational transcripts and collecting a new dataset, 
\system\footnote{We release the collected \system dataset at: \href{https://github.com/huashen218/MultiTurnCleanup.git}{https://github.com/huashen218/MultiTurnCleanup.git}}.
We design a data labeling schema to collect the high-quality dataset and provide extensive data analysis. 
Furthermore, we leverage two modeling approaches for experimental evaluation as benchmarks for future research.
\end{abstract}
















\section{Introduction}
\label{sec:introduction}

\begin{figure}[!t]
    \centering
    \includegraphics[width=1.\columnwidth]{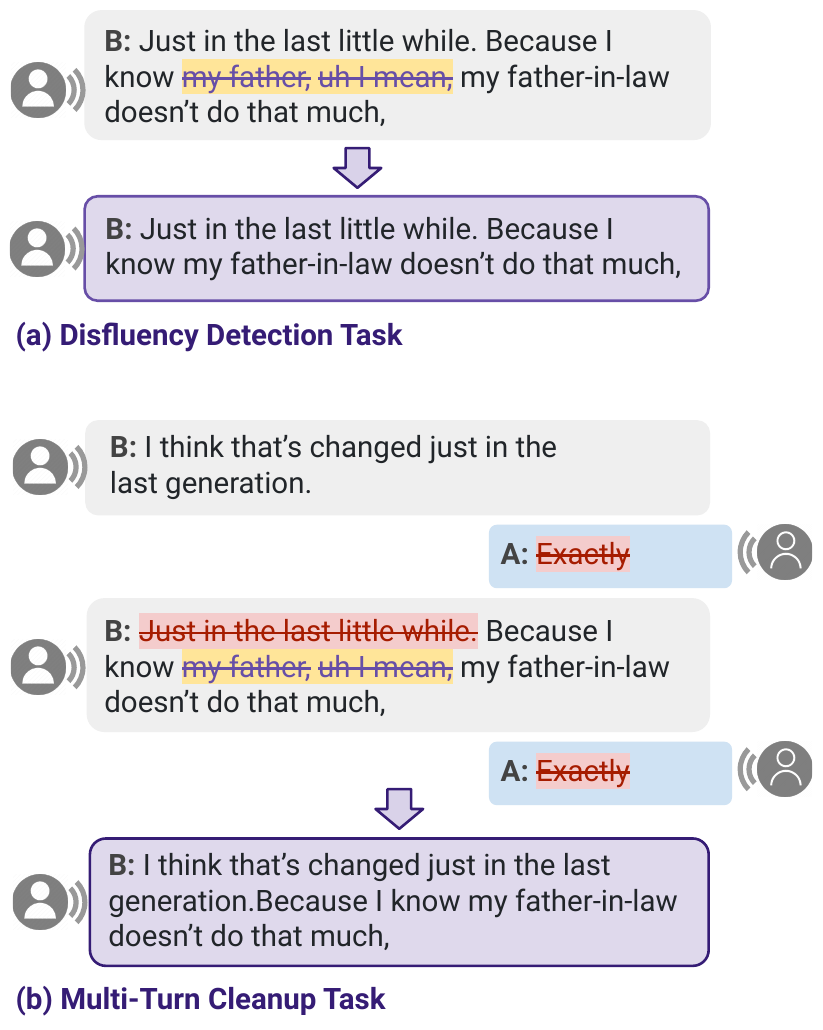}
    \vspace{-.5pc}
    \caption{A comparison of (a) existing Disfluency Detection Task (yellow highlights indicate disfluencies) with (b) the proposed \task (red highlights indicate multi-turn cleanups) for spoken conversational transcripts.
    }
    \label{fig:teaser}
 \end{figure}

\begin{table*}[!t]
\centering
\footnotesize
\begin{tabular}[t]{@{} p{0.15\textwidth} | p{0.23\textwidth} | p{0.1\textwidth} | p{0.42\textwidth} @{}}
\toprule
\textbf{Category}  & \textbf{Definition} &  \multicolumn{1}{l|}{\textbf{Count (\%)}} & \textbf{Conversation Instance} \\
\midrule
\color[HTML]{1C5CCC}{\colorbox{ack-bc}{Acknowledgment} \colorbox{ack-bc}{   and Confirmation}} & Speakers show that they are listening to and agree with the other speakers  & 24.3k (17\%) &  \multirow{5}{*}{\begin{tabular}[t]{@{}l@{}} 
\textbf{A}: I guess both of us are very much aware of the \\
equality. it seems like women are, just starting to \\
get kind of equality in jobs and the home where \\
husbands are starting to doing dishes, {\color[HTML]{38761D} \colorbox{inc-bc}{\st{or some}}} \\
\textbf{B}: I think that's changed just in the last generation. \\
\textbf{A}: \color[HTML]{1C5CCC}{\colorbox{ack-bc}{\st{Exactly.}}} \\
\textbf{B}: {\color[HTML]{A61C00}\colorbox{rep-bc}{\st{Just in the last little while.}}} Because \\
I know my father-in-law doesn't do that much, \\
\textbf{A}: \color[HTML]{1C5CCC}{\colorbox{ack-bc}{\st{Exactly.}}} \\
\textbf{B}: of dishes, taking care of kids, {\color[HTML]{B45F06} \colorbox{think-bc}{\st{or what else,}}} \\ {\color[HTML]{B45F06}\colorbox{think-bc}{\st{you know,}}} that kind of stuff but my husband is \\ wonderful. \\
\textbf{A}: that's the way my husband is too. it doesn't \\
bother him to do the dishes, {\color[HTML]{A61C00}\colorbox{rep-bc}{\st{it doesn't bother}}} \\ 
{\color[HTML]{A61C00}\colorbox{rep-bc}{\st{him to do}}} the laundry verses, men from way back\\
, {\color[HTML]{351C75}\colorbox{other-bc}{\st{there is that,}}} if you did that you were henpecked.
\end{tabular}}  \\
\cmidrule{1-3}
\color[HTML]{A61C00}{\colorbox{rep-bc}{Repetition and} \colorbox{rep-bc}{Paraphrase}}       & {\color[HTML]{212529} Speakers may repeat or paraphrase their words during the conversation.}  & 30k (21\%) &  \\
\cmidrule{1-3}
{\color[HTML]{B45F06} \colorbox{think-bc}{Think aloud}}                     & {\color[HTML]{212529} Speakers talk to themselves during thinking instead of talking to others.}    & 15.7k (11\%)  &\\
\cmidrule{1-3}
{\color[HTML]{38761D} \colorbox{inc-bc}{Incomplete} \colorbox{inc-bc}{Sentences}}   & {\color[HTML]{212529} Speakers may also say incomplete sentences due to interruption, changing topics, etc.} &  47.2k (33\%)  & \\
\cmidrule{1-3}
{\color[HTML]{351C75} \colorbox{other-bc}{Others}}   &  The remaining discontinuity categories.  & 25.8k (18\%) & \\
\bottomrule
\end{tabular}
\vspace{-.5pc}
\caption{The linguistic taxonomy (\textbf{Category}) and definition (\textbf{Definition}) of discontinuities in the \system dataset for the \task. We further provide the statistics of each category (\textbf{Count(\%)}) in the dataset and a conversational instance (\textbf{Conversation Instance}), where \textbf{A} and \textbf{B} indicate two speakers.}

\label{tab:category}
\vspace{-1.pc}
\end{table*}



%
%
Spontaneous spoken conversations contain interruptions such as filled pauses, self-repairs, etc. \cite{shriberg1994preliminaries}.
%
These phenomena act as noise that hampers human readability~\cite{adda2003disfluency} and the performance of downstream tasks such as question answering~\cite{gupta2021disfl} or machine translation~\cite{hassan2014segmentation}
on transcripts of human spoken conversations.
State-of-the-art disfluency detection methods~\cite{yang2020planning,lou2020improving} identify and remove disfluencies in order to improve the readability of spoken conversational transcripts~\cite{wang-etal-2022-adaptive,chen-etal-2022-teaching}.
For instance, Figure~\ref{fig:teaser}(a) shows that disfluency detection methods can remove self-repairs of single turns.
However, these models focus on removing interruptions and errors that commonly occur within single-turn utterances and cannot handle discontinuities across multiple turns.
%
%
For example, in Figure~\ref{fig:teaser}(b), speaker \spkb is in the middle of a thought when Speaker \spka interrupts to signal that they are following along (``\spka: Exactly''). \spkb continues their train of thought (``\spkb: Just in the last little while. Because...'') by paraphrasing their own last sentence (``...just in the last generation.''). 
The result is an exchange that is longer and more difficult to follow than necessary to understand what \spkb is conveying. 
%
%

%
%

This paper aims to ``clean up" spoken conversation transcripts by detecting these types of multi-turn ``discontinuities" inherent in spontaneous spoken conversations.
Once detected, they can be removed to produce transcripts that look more like hand-written conversations conducted over text messaging, social media, or e-mail as shown in Figure \ref{fig:teaser}(b).
Given that this is a novel task, with no pre-existing labeled data or benchmarks, we first define a taxonomy of non-disfluency discontinuities (see Figure \ref{tab:category}). Then we collect a new dataset, \system, for the \textbf{Multi Turn} spoken conversational transcript \textbf{Cleanup} task, based on the Switchboard Corpus~\cite{godfrey1992switchboard} and label according to the proposed taxonomy.
Finally we develop two baseline models to detect these discontinuities which we evaluate as benchmarks for future \task studies.
Our data analysis suggests that the \system dataset is of high quality. We believe it will help to facilitate research in this under-investigated area. 
\section{Data Collection and Analysis}
\label{sec:analysis}

We propose an innovative \task and collect a novel dataset for this task called \system\footnote{The data collection cost was about $\$18,000$, with payment as $\$0.40$/HIT ($\$8$/hour assuming 3min/HIT) and bonuses ($\$2-\$100$/worker) for top raters. }. This section presents the task definition, data collection process, and analysis.

%
\subsection{Task Definition}

Compared with the existing \emph{disfluency detection} task, which aims to detect disfluencies (\eg self-repairs, repetitions, restarts, and filled pauses) that commonly occur within single-turn utterances~\cite{rocholl2021disfluency,chen-etal-2022-teaching}, the \textbf{\task} requires identifying discontinuities both within a single turn and across multiple turns in the multi-party spoken conversational transcripts.
To explicitly define the task and discontinuity taxonomy, we conducted an in-depth analysis of the Switchboard corpus\footnote{Switchboard corpus is a collection of five-minute human-human telephone conversations. We chose Switchboard for data construction because: 1) it has relatively large size; and 2) it contains ground-truth disfluency annotations that our study can build upon to annotate multi-turn cleanup labels.}~\cite{godfrey1992switchboard}.
Specifically, we randomly sampled a subset of Switchboard conversations, annotated the discontinuity spans other than existing disfluency types, and grouped the annotated discontinuities into five main categories. Note that we conducted the discontinuity annotation and category grouping process iteratively with all authors to reach the consensus. We demonstrate the finalized taxonomy of discontinuities in Table~\ref{tab:category}.

\begin{figure*}[!t]
    \centering
    \includegraphics[width=1.\textwidth]{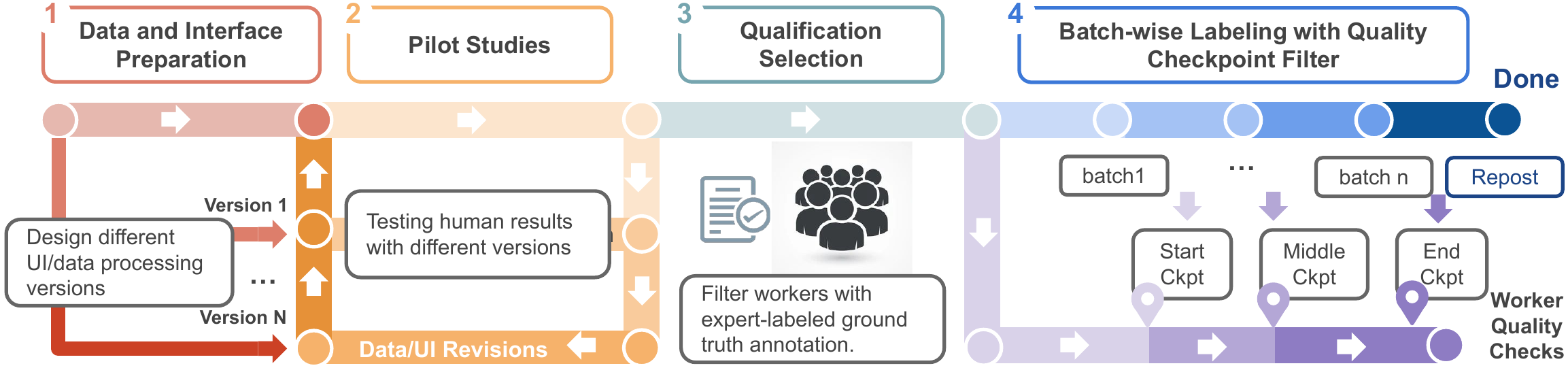}
    \vspace{-.5pc}
    \caption{A four-step data labeling schema to collect \system dataset for the \task on spoken conversational transcripts. This schema enabled efficient collection of the high-quality \system dataset via MTurk platform, where annotators labeled cleanup marks and corresponding categories.}
    \label{fig:schema}
 \end{figure*}

\subsection{Data Preprocessing}
%

We preprocessed the Switchboard corpus by automatically removing the single-turn disfluencies with pre-defined rules based on Treebank-3~\cite{marcus1999treebank}\footnote{See Appendix~\ref{sec:preprocessing} for detailed preprocessing rules and data statistics.}. As a result, we could encourage our recruited humans annotators to concentrate on cleaning up multi-turn discontinuity phenomena based on the five categories in Table~\ref{tab:category}.
%
%


%
%
We split each conversation into multiple chunks, where each chunk composes one HIT (Human Intelligence Task) containing around 300 tokens. We further ensure that the successive chunks overlap around 50\% of tokens, providing enough context for each conversation fragment. The resulting data preprocessing statistics are shown in Table~\ref{tab:data_preprocessing} in Appendix~\ref{sec:preprocessing}. 

\begin{table}[!t]
\footnotesize
\setlength{\tabcolsep}{6pt}
\begin{tabular}{cccccc}
\toprule
\cellcolor[HTML]{DAE8FC}\textbf{Datasets} & \cellcolor[HTML]{DAE8FC}\textbf{\#Conv} & \cellcolor[HTML]{DAE8FC}\textbf{\#Turns} & \cellcolor[HTML]{DAE8FC}\textbf{\#Tokens} & \cellcolor[HTML]{DAE8FC}\textbf{\#Cleanup} \\
\midrule
\textbf{Train} & 932 & 74k  & 1M  & 132k \\
\textbf{Dev}   & 86  & 3.7k & 60k & 6.1k   \\
\textbf{Test}  & 64  & 2.9k & 43k & 5k   \\
\midrule
\textbf{Sum} & \textbf{1082}           & \textbf{81k}   & \textbf{1.1M} & \textbf{143k} \\
\bottomrule
\end{tabular}
\vspace{-.5pc}
\caption{Statistics of \system dataset.}
\label{tab:data_statistics}
\vspace{-.7pc}
\end{table}

\subsection{Labeling Procedure}
%
%
%

Given the preprocessed data, we then conducted the human annotation process based on a data labeling schema\footnote{Steps 3 and 4 lasted about one month. More annotation quality control details are available in Appendix~\ref{sec:keynotes}.} shown in Figure~\ref{fig:schema}.

%

\noindent\textbf{Preparation and qualification selection.} 
In steps 1 and 2, we prepared a suite of data preprocessing and user interface (UI) variations and conducted seven pilot studies to select the optimal task design. The final UI (see Appendix~\ref{sec:ui}) consists of: \emph{i)} an introduction to the task, \emph{ii)} an annotation example with highlighted discontinuities, and \emph{iii)} the task workspace with affordances for annotation. 
In step 3, we recruited a set of qualified MTurk workers using a ``Qualification HIT''. 
We compared all 580 workers' submissions with the ground truth (authors' consensus) and select the 222 workers (38.3$\%$) with an F1 $\geq$ 0.3\footnote{The $0.3$ threshold is reasonable due to task subjectivity.} to participate in step 4. 
%

\vspace{2pt}
\noindent\textbf{Large-scale data labeling.}
Controlling annotation quality for large-scale data labeling is challenging in MTurk~\cite{daniel2018quality}.
To address this, we employed a ``batch-wise labeling with quality checkpoint filter'' \cite{bragg2016optimal}.
Specifically, we split the dataset into small batches and posted them with ``Quality Checkpoint HITs'' (QCH) mixed in. 
Overall, we posted 22 batches including 7277 HITs and 11 QCH in total.
We leverage these checkpoint HITs to exclude unqualified workers (F1 $\leq$ 0.3).

\noindent\textbf{Annotation filtering and aggregation.}
After finishing the final batch, we collected all annotated batches and excluded 72
unqualified workers with all their HITs.
Then we reposted 26\% 
of the assignments where the conversations had less than two annotations to the remaining qualified workers.
Finally, we aggregated the annotations for each turn by only keeping the best worker's (highest F1-score) labels to compose the \system dataset.
The average F1 for raters of labeled turns in \system is 0.57.
We summarize the
per-category statistics in Table~\ref{tab:category} and
the statistics of \system in Table~\ref{tab:data_statistics}\footnote{We leave out sw4[2-4]* subgroups in Switchboard as they are less commonly used, resulting in 1082 total conversations.}.

\begin{table}[!t]
\footnotesize
\setlength{\tabcolsep}{6.5pt}
\begin{tabular}{lccccc}
\toprule
{\cellcolor[HTML]{DAE8FC}\textbf{IRR}}  & {\cellcolor[HTML]{DAE8FC} \textbf{Experts}}  & \multicolumn{4}{c}{\cellcolor[HTML]{DAE8FC}{\textbf{MTurk Workers}}} \\
\midrule
 &  & \textbf{Train}        & \textbf{Dev}        & \textbf{Test}        & \textbf{All}        \\
 \cmidrule{3-6}
\multirow{-2}{*}{\cellcolor[HTML]{FFFFFF}\textbf{\makecell{Fleiss' \\ Kappa}}} & \multirow{-2}{*}{\cellcolor[HTML]{FFFFFF} 0.596} & 0.560 & 0.592  & 0.557 & 0.561  \\ \bottomrule
\end{tabular}
\vspace{-.5pc}
\caption{The averaged Fleiss' Kappa scores of all conversation turns in the \system dataset.}
\label{tab:fleiss_kappa}
\end{table}

 \begin{figure*}[!t]
    \centering
    \includegraphics[width=\textwidth]{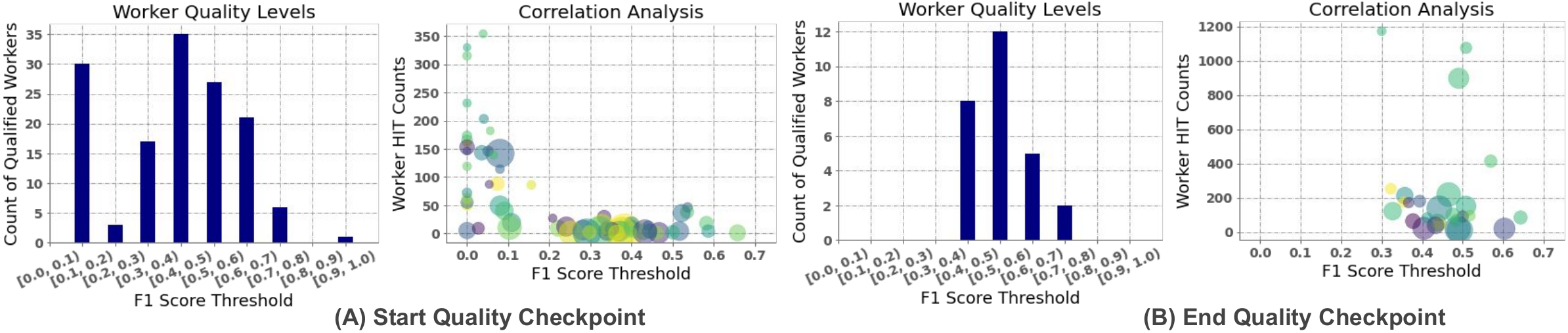}
    \vspace{-1.3pc}
    \caption{The comparison of worker quality performance between (A) start checkpoint and (B) end checkpoints. For each one, we plot participated workers' F1 score distribution (left) and the correlation between each worker's F1 score and finished HIT count (right), the circle size means each worker's averaged elapsed time to finish a HIT.
    }
    \label{fig:worker_quality}
\end{figure*}

\subsection{Validating Human Annotation Accuracy}
\label{sec:accuracy}

During the whole data labeling process, we consistently assessed the human annotation accuracy and filtered out unqualified workers to control the data quality. 
We visualize the annotation quality in terms of the distribution of workers' F1 scores (see Figure~\ref{fig:worker_quality}(A)(B)-left), as well as the correlation between each worker's F1 score and their finished HIT counts and the average elapsed time per HIT (see Figure~\ref{fig:worker_quality}(A)(B)-right).
These figures show how removing unqualified annotations with checkpoints can effectively control quality during the annotation process.
%
Particularly, we observe that at the start (A), even after passed our initial ``Qualification HIT'' in step 3, 23\% of workers perform at F1 $<$ 0.3 but complete over $80\%$ of all assignments, leaving only a limited amount of data for more competent workers to label.
By continually excluding unqualified
workers with F1 $<$ 0.3, all remaining workers have F1 $\geq$ 0.3 by the final batch (B). 

\subsection{Turn-based Inter-Rater Reliability}

We compute Inter-Rater Reliability using Fleiss' Kappa Agreement~\cite{fleiss1973equivalence} for each annotated turn and average all turns' scores. Table~\ref{tab:fleiss_kappa} shows that the workers' Fleiss' Kappa scores are comparable to those of the authors.

\section{Multi-turn Cleanup Models}
\label{sec:model}

\begin{table}[!t]
\footnotesize
\setlength{\tabcolsep}{8.4pt}
\begin{tabular}[t]{cllll}
\toprule
\cellcolor[HTML]{DAE8FC}\textbf{Sub-tasks}      & \cellcolor[HTML]{DAE8FC}\textbf{Model}  & \multicolumn{1}{c}{\cellcolor[HTML]{DAE8FC}\textbf{F1}} & \multicolumn{1}{c}{\cellcolor[HTML]{DAE8FC}\textbf{R}} & \multicolumn{1}{c}{\cellcolor[HTML]{DAE8FC}\textbf{P}} \\
\midrule
\makecell{\textsc{Disfluency}}  &\makecell{\std}  & 89.8 &   88.3     &   91.2   \\
\midrule
\multirow{2}{*}{\makecell{\textsc{MultiTurn} \\ \textsc{Cleanup}}} & \makecell{\textsc{Baseline}} &   15.5      &   8.77 & 65.8  \\
\cmidrule{2-5}
 & \makecell{\mtd} &   56.8      &   55.4 & 58.3  \\
\midrule
\end{tabular}
\vspace{-.7pc}
\caption{Model performance on the two sub-tasks, including detecting single-turn disfluencies with \textsc{Disfluency} dataset and multi-turn discontinuities with the proposed \system dataset.
}
\label{tab:model_two_stage}
\vspace{-1.pc}
\end{table}

Given the collected \system dataset, we leverage two different BERT-based modeling approaches, including a two-stage model and a combined model, for the \task to remove both single-turn disfluencies and multi-turn discontinuities.


\subsection{The Two-Stage Model}
%
%
The two-stage model is composed of a Single-Turn Detector (\std) to remove the traditional single-turn disfluencies and a successive Multi-Turn Detector (\mtd) to remove the discontinuities occurring across multiple turns.
We employ the BERT-based modeling, presented in \citet{rocholl2021disfluency}, for both \std and \mtd stages.
Particularly, we fine-tune the \std based on the traditional single-turn disfluency dataset~\cite{godfrey1992switchboard}, whereas the \mtd is fine-tuned based on our collected \system dataset.
We concatenate \std and \mtd successively into the pipeline of the two-stage model, so that both the single-turn disfluencies and multi-turn discontinuities in the raw conversational transcript can be removed with one pass.

\subsection{The Combined Model}
%
We design the combined model, using only one BERT-based detector~\cite{rocholl2021disfluency}, to simultaneously remove both single-turn disfluencies and multi-turn discontinuities.
To this end, we create a \textsc{UnionDiscontinuity} dataset, which combines both the single-turn disfluency and multi-turn discontinuities labels in \citet{godfrey1992switchboard} and \system datasets, respectively.
Then we achieve the combined model by fine-tuning the detector with this \textsc{UnionDiscontinuity} dataset.

\section{Experiments}
\label{sec:evaluation}

\begin{table}[!t]
\footnotesize
\setlength{\tabcolsep}{9pt}
\begin{tabular}[t]{cclll}
\toprule
\cellcolor[HTML]{DAE8FC}\textbf{}  & \cellcolor[HTML]{DAE8FC}\textbf{Model}  & \multicolumn{1}{c}{\cellcolor[HTML]{DAE8FC}\textbf{F1}} & \multicolumn{1}{c}{\cellcolor[HTML]{DAE8FC}\textbf{R}} & \multicolumn{1}{c}{\cellcolor[HTML]{DAE8FC}\textbf{P}} \\
\midrule
\multirow{3}{*}{\makecell{\textbf{Multi-Turn} \\ \textbf{Cleanup} \\ \textbf{Task} }} 
& \makecell{\textsc{Baseline}} & 58.2    &  42.5    &  92.3 \\
\cmidrule{2-5}
& \makecell{Two-Stage} & 68.2    &   64.6   & 72.3 \\
\cmidrule{2-5}
& \makecell{Combined} & 74.9   &  72.9 &  76.9    \\
\toprule
\end{tabular}
\vspace{-.7pc}
\caption{Modle performance on the overall \task with the \textsc{UnionDiscontinuity} dataset, where the Combined Model achieves the best F1 score. 
}
\label{tab:model_combined}
\vspace{-1.pc}
\end{table}

\subsection{Experimental Setup}

\noindent\textbf{The Two-Stage Model}. The \std and \mtd are separately trained. We train the \std with the existing disfluency dataset, where the input is a single sentence (\ie slash unit), with a maximum sequence length of 64. In comparison, we train the \mtd with \system dataset, where the input consists of multiple slash-units (demarcated with [SEP] token between turns) with a maximum sequence length of 512. We feed full transcripts to the \mtd in chunks with an overlap of 50\% for prediction context. Then we predict discontinuities where either of the overlapping predictions for a given token was positive.
During inference, the stage-2 \mtd module loads the outputs from the stage-1 \std module, removes all of the tokens classified as disfluencies, and uses this redacted texts as its own input.

%



\noindent\textbf{The Combined Model}. We train the combined model with the \textsc{UnionDiscontinuity} dataset using the same training settings of the aforementioned \mtd module. 
During inference, we predict both single-turn and multi-turn discontinuities, as nondistinctive labels, simultaneously.

\noindent\textbf{Baseline}. We employ the state-of-the-art BERT based disfluency detection model~\cite{rocholl2021disfluency} trained with the widely used disfluency dataset~\cite{godfrey1992switchboard} as the \textsc{Baseline}.

\noindent\textbf{Deployment}.
We train the models on Google's AutoML platform, where it selects the optimal training settings as: Adam optimizer with learning rate as $1e-5$, batch size of 8, and  1 epoch.


\subsection{Evaluation Metrics}
%
%
We evaluate all models' performance with per-token Precision (\textbf{P}), Recall (\textbf{R}), and F1 score (\textbf{F1}) on predicting if the token should be cleaned as single-turn disfluencies (\std of two-stage model), or multi-turn discontinuities (\mtd of two-stage model), or their mixtures (the combined model). 
%
%
%

\subsection{Results}



\noindent\textbf{Evaluation on two sub-tasks}.
The \task inherently involves two different sub-tasks, including the single-turn disfluency detection (\ie with \textsc{Disfluency} dataset) and multi-turn discontinuity detection (\ie with our collected \system dataset), we first validate that the presented models can achieve state-of-the-art performance on the two sub-tasks (\ie with the two different datasets), respectively.

Particularly, Table~\ref{tab:model_two_stage}
illustrates the performance of \textsc{Baseline} and presented models on the two datasets. 
The \std module achieves cutting-edge performance~\cite{chen-etal-2022-teaching} to detect single-turn disfluencies. Also, the \mtd module outperforms the \textsc{Baseline} on detecting multi-turn discontinuities with our proposed \system dataset. 
The significant disparity between \mtd and \textsc{Baseline} methods (\eg 56.8 vs. 15.5 in F1) also indicate the difficulty of detecting multi-turn discontinuities in \system dataset.
%



\noindent\textbf{Evaluation on removing all discontinuities}.
Furthermore, we evaluate the overall model performance on jointly detecting the single-turn disfluencies and multi-turn discontinuities with one pass based on the \textsc{UnionDiscontinuity} dataset. 
As shown in Table~\ref{tab:model_combined}, we observe that both the proposed Two-Stage Model and Combined Model can outperform \textsc{Baseline} method. In addition, the Combined Model achieves a 6.7 higher F1 score than Two-Stage Model on the \task.

\vspace{-0.5pc}
\section{Related Work}
\vspace{-0.5pc}
\label{sec:literature}


Recent disfluency detection studies develop BERT-based models~\cite{bach2019noisy,rocholl2021disfluency,rohanian2021best} and show significant improvement over LSTM-based models~\cite{zayats2016disfluency,wang2016neural,hough2017joint} in disfluency detection tasks.
%
Prior studies also show the importance of data augmentation methods, by leveraging extra transcript sources to improve disfluency detection performance~\cite{lou2018disfluency,lou2020improving}.
%
While most of the research has been focused on improving single-turn disfluency detection accuracy,
little exploration has been done in detecting multi-turn transcript discontinuities. 

Obtaining reliable annotated datasets via crowdsourcing is challenging and expensive~\cite{alonso2014crowdsourcing,wong2022ground,northcutt2021pervasive}.
%
%
%
 To collect qualified dataset for multi-turn cleanup task, this work designs a data labeling schema which efficiently collects qualified dataset via the MTurk.


\vspace{-0.5pc}
\section{Limitation and Conclusion}
\vspace{-0.5pc}
\label{sec:conclusion}


We are aware that, in some specific scenarios, it might be undesirable to remove some multi-turn discontinuities because they convey social meaning in human interactions (\eg engagement). We address this issue by providing category labels. As a result, future research can flexibly select subsets of the \system labels to train the model and clean up multi-turn discontinuities.
%

%

%
%
This study defines an innovative Multi-Turn Cleanup task and collects a high-quality dataset for this task, named \system, using our presented data labeling schema. We further leverage two modeling approaches for experimental evaluation as the benchmarks for future research.
%
%


\section*{Acknowledgements}
We thank Noah B. Murad for his help in conducting the human evaluation experiments. We also thank Shyam Upadhyay, Daniel J. Liebling, Tiffany Knearem, Kenton Lee, and other Google colleagues for providing their constructive feedback on this study.
We thank the Amazon MTurk workers for their excellent annotations.
We thank the reviewers for their thoughtful comments.

\section*{Ethics Statement}
%
%


The collected \system dataset is built upon the published Switchboard Corpus~\cite{godfrey1992switchboard}. The dataset is sufficiently anonymized, so it is impossible to identify individuals.
In addition, we protect privacy during the data collection process through the MTurk platform. The posted dataset does not list any identifying information of MTurk workers. Also, the data collection process does not access any demographic or confidential information (\eg identification, gender, race, etc.) from the MTurk workers.
In general, the dataset can be safely used with low risk in research and application fields for cleaning up spoken conversations and speech transcripts.

\bibliography{custom}
\bibliographystyle{acl_natbib}

\appendix

\section{Appendix}
\label{sec:appendix}
\subsection{Definitions of Disfluency Detection and Annotations}
\label{sec:disfluency}

\noindent\textbf{Disfluency Definition.} When humans speak, our language is peppered with interruptions and errors known as disfluencies. Formally, disfluencies are irregularities that are an integral part of spontaneous speech and include \emph{self-repairs}, \emph{repetitions}, \emph{restarts}, and \emph{filled pauses}~\cite{schegloff1977preference}.

\paragraph{\textbf{Disfluency Annotations.}} Following~\citet{shriberg1997prosody}, the disfluency annotation includes:
\begin{itemize}
    \item the \emph{\textbf{reparandum}}: the material that the speaker intends to delete.
    \item the \emph{\textbf{interruption point}}: denoted as (+).
    \item optional \emph{\textbf{interregnum}}: enclosed in curly brackets, which include \emph{filled pauses} and \emph{discourse markers}, such as ``uh'', ``um'', ``you know'', ``I mean'', etc.
    \item optional \emph{\textbf{repair}}: the material that semantically replaces the reparandum.
\end{itemize}

\noindent Some examples of disfluency annotation include:





\begin{itemize}
    \item \small{\texttt{
$[$ it's + \{ uh \} it's $]$ almost ...
}}
    \item \small{\texttt{
$[$ was it, + \{ I mean, \} did you $]$ put...
}}
    \item \small{\texttt{
$[$ I just + I $]$ enjoy working
}}
    \item \small{\texttt{
$[$ By + $]$ it was attached to ...
}}
\end{itemize}

\subsection{Dataset Preprocessing Details.}
\label{sec:preprocessing}

\begin{table}[ht]
\footnotesize
\setlength{\tabcolsep}{7pt}
\begin{tabular}{llllll}
\toprule
\cellcolor[HTML]{DAE8FC}\textbf{Datasets} & \cellcolor[HTML]{DAE8FC}\textbf{\#Conv} & \cellcolor[HTML]{DAE8FC}\textbf{\#HITs} & \cellcolor[HTML]{DAE8FC}\textbf{\#Turns} & \cellcolor[HTML]{DAE8FC}\textbf{\#Tokens} \\
\midrule
\textbf{Train} & 932                     & 6579            & 142,135          & 1,928,169 \\
\textbf{Dev} & 86                      & 396             & 7,250            & 115,756 \\
\textbf{Test} & 64                      & 302             & 5,888            & 85,321 \\
\midrule
\textbf{Sum} & \textbf{1082}           & \textbf{7277}   & \textbf{155,273} & \textbf{2,129,246} \\
\bottomrule
\end{tabular}
\caption{Statistics of data preprocessing results.}
\label{tab:data_preprocessing}
\end{table}

We preprocess the Switchboard corpus~\cite{godfrey1992switchboard} with the human-annotated disfluencies based on Treebank-3~\cite{marcus1999treebank}.

Based on the human annotations, we first cleaned the sentences by removing the non-speech events and words/phrases with the markers including:

\begin{itemize}
    \item Prosodic markup, like $\#$ and $/$.
    \item Other language markup, like <English bike>.
    \item Noise markers, like <<laughter>> or <Throat\_clearing>
    \item Noise markers with curly brackets, like \{breathing\} or \{gasp\} or \{lipsmack\} or \{again, imitates the sound of whales\}
    \item Double parentheses from uncertainty markers, like ((yesterday))
    \item Plus signs surrounding words, like +sight-seeing+.
    \item Context markers, like (laughter) or (RECESS).
    \item Punctuations, like $\%$, **, $\&$, />, +>, <]>, (), ((, )), [[, ]].
\end{itemize}

In the next stage, we conservatively remove the following types of disfluencies recursively:

\begin{itemize}
    \item Reparandum -- the material that the speakers intends to delete.
    \item Interregnum -- including types of:
    \begin{itemize}
        \item Discourse markers (\ie marked with $\{\textbf{D}...\}$ like \emph{you know}, \emph{well}, \emph{so}, \emph{like}, etc.).
        \item Explicit editing term (\ie marked with $\{\textbf{E}...\}$ like \emph{I mean}, \emph{sorry}, \emph{excuse me}, etc).
        \item Filler words (\ie marked with $\{\textbf{F}...\}$ like \emph{uh}, \emph{um}, \emph{huh}, \emph{oh}, etc.
    \end{itemize}
\end{itemize}

By removing all the markers and disfluencies described above, we present the remaining contents of Switchboard corpus to MTurk workers for labeling the \system dataset for \task.

\begin{figure*}[!t]
    \centering
    \includegraphics[width=0.77\textwidth]{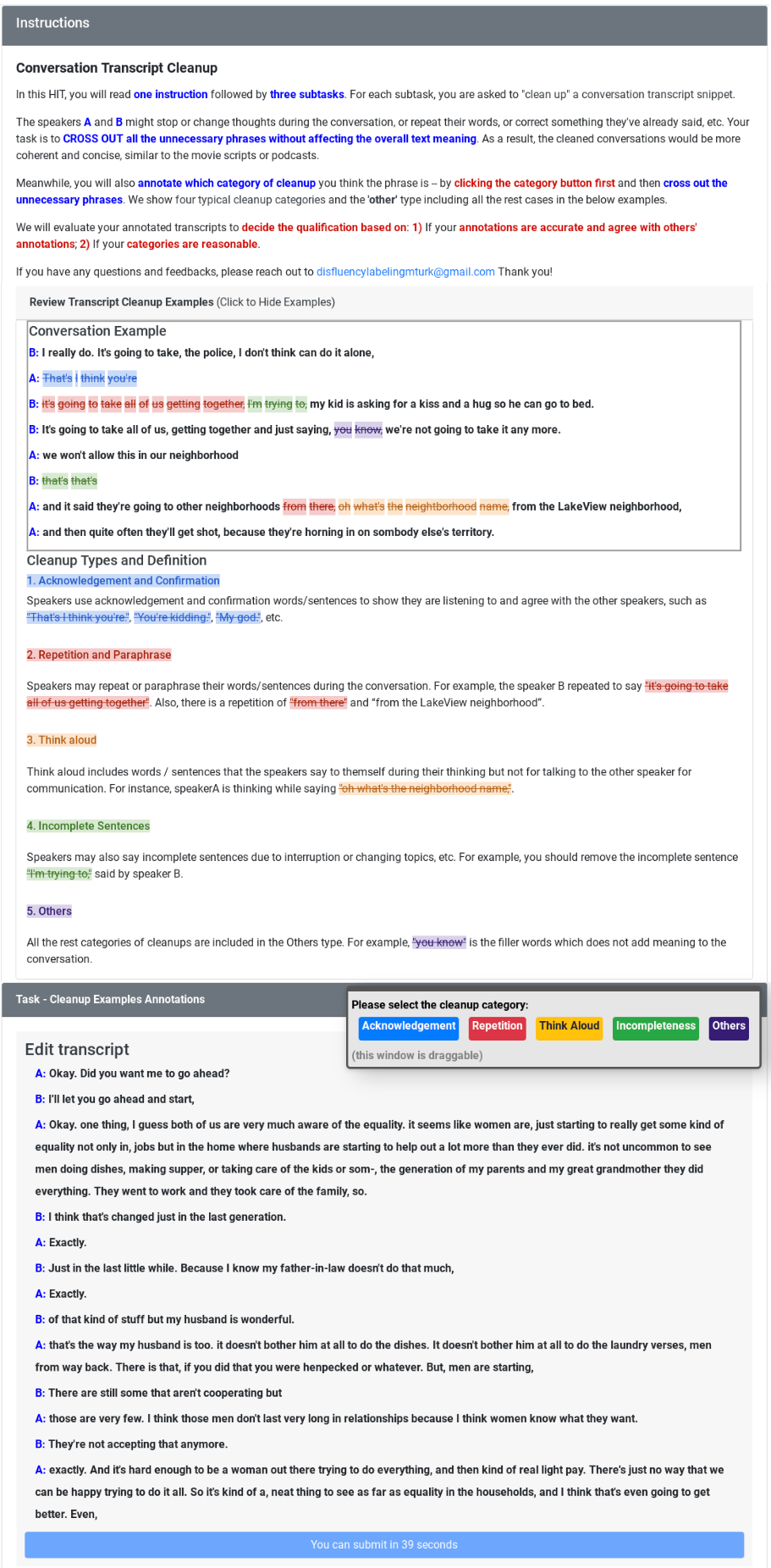}
    \caption{An example of the User Interface for the MTurk worker annotation. The ``Review Transcript Cleanup Example'' section is foldable by clicking to hide and show the example.}
    \label{fig:ui}
 \end{figure*}



\subsection{Key notes of data labeling.}
\label{sec:keynotes}

We notice a list of key points that are imperative to ensure the data quality and avoid spammers in MTurk platform. 
%
\begin{itemize}
    \item \emph{Keep task simple and instructions clear.} This helps workers to better understand the task, reduce their cognitive load and focus on annotating qualified outputs.
    \item \emph{Select workers and check quality constantly.} Repeatedly inspecting worker quality can significantly reduce spammers and improve data quality, as more details validated in Sec~\ref{sec:accuracy}.
    \item \emph{Communicate via emails.} Notifying new post or notes to workers helps improve return rate significantly. Also, their feedback may be valuable to improve the data labeling.
\end{itemize}


\subsection{User Interface Demo}
\label{sec:ui}

We demonstrate the finalized User Interface (UI) design of the HIT for the large-scale datasets in Figure~\ref{fig:ui}.



\end{document}